\documentclass[letterpaper, 10 pt, conference]{ieeeconf}  % Comment this line out if you need a4paper

\IEEEoverridecommandlockouts                              % This command is only needed if 
\usepackage{graphicx}
\usepackage{subfigure}
\usepackage{amsmath} % assumes amsmath package installed
\usepackage{amssymb}  % assumes amsmath package installed
\usepackage{bm}

\title{\LARGE \bf
Visual Explanation of Deep Q-Network for Robot Navigation by Fine-tuning Attention Branch
}

\author{Yuya Maruyama$^{1}$, Hiroshi Fukui$^{1}$, Tsubasa Hirakawa$^{1}$, Takayoshi Yamashita$^{1}$,\\Hironobu Fujiyoshi$^{1}$, and Komei Sugiura$^{2}$%
\thanks{$^{1}$Y. Maruyama, H. Fukui, T. Hirakawa, T. Yamashita and H. Fujiyoshi are with Chubu University, 1200 Matsumoto, Kasugai, Aichi, Japan
        {\tt\small \{maru94, fhiro, hirakawa\}@mprg.cs.chubu.ac.jp, \{takayoshi, fujiyoshi\}@isc.chubu.ac.jp}}%
\thanks{$^{2}$Komei Sugiura is with Keio University, 3--14--1 Hiyoshi, Kohoku, Yokohama, Kanagawa, Japan
        {\tt\small komei.sugiura@keio.jp}}%
}

\begin{document}

\maketitle
\thispagestyle{empty}
\pagestyle{empty}

%%%%%%%%%%%%%%%%%%%%%%%%%%%%%%%%%%%%%%%%%%%%%%%%%%%%%%%%%%%%%%%%%%%%%%%%%%%%%%%%
\begin{abstract}

Robot navigation with deep reinforcement learning (RL) achieves higher performance and performs well under complex environment. Meanwhile, the interpretation of the decision-making of deep RL models becomes a critical problem for more safety and reliability of autonomous robots. In this paper, we propose a visual explanation method based on an attention branch for deep RL models. We connect attention branch with pre-trained deep RL model and the attention branch is trained by using the selected action by the trained deep RL model as a correct label in a supervised learning manner. Because the attention branch is trained to output the same result as the deep RL model, the obtained attention maps are corresponding to the agent action with higher interpretability. Experimental results with robot navigation task show that the proposed method can generate interpretable attention maps for a visual explanation.

\end{abstract}

%===========================================================
\section{INTRODUCTION}
\label{sec:intro}

Deep reinforcement learning (RL) has been achieved great performances in various applications such as video game \cite{Mnih2013,Mnih2015}, Go \cite{Silver2016,Silver2017}, and robotics applications \cite{Fangyi2015,Smolyanskiy2017,Zhu2017,Tai2017,Chen2017}.
One of the major application of deep RL for robotics is an autonomous robot navigation \cite{Fangyi2015,Smolyanskiy2017,Zhu2017,Tai2017}.
Deep RL models have a massive number of parameters in a network and those parameters are updated by training via trial-and-error.
Consequently, deep RL models can successfully avoid obstacles and move under a complex environment in which obstacles or pedestrians exist.
While deep RL achieves higher performances on various tasks, deep learning-based approaches including deep RL have a difficulty that is hard to interpret the deep RL agent's decision-making.

Toward the interpretation of deep neural network's decision-making, \textit{visual explanation} has been widely studied in the computer vision field \cite{Zeiler2014,Zhou2016,Smilkov2017,Selvaraju2017,Fukui2018}.
In visual explanation problem, we generate a map that highlights regions where a convolutional neural network (CNN) focuses on.
The map is called an \textit{attention map}.
Then, we visually interpret the reason for the network output from the obtained attention map.
% top-down and bottom-up の説明
Visual explanation methods can be categorized into two approaches; bottom-up and top-down.
Bottom-up visual explanation approach calculates an attention map by using response values of convolutions.
Meanwhile, a top-down visual explanation approach generates an attention map based on network output, that is, the categories of a classification problem.
The bottom-up approaches have been used for various deep learning methods because bottom-up methods can be easily applied for any pre-trained networks.
Although top-down approaches need to construct a network to output attention maps during inference, top-down methods can output attention maps while obtaining the network output.

Visual explanations are applied not only computer vision tasks but also deep RL-based tasks \cite{Sorokin2015,Pardo2018,Samuel2018}.
Visual explanations on deep RL use bottom-up approaches and top-down approaches have not been used for deep RL tasks.
However, a top-down visual explanation approach plays an important role in deep RL-based robot navigation tasks existing human-robot interactions.
If the robot provides the reason for decision-making on taking an action, the surrounding human can understand the robot behaviors.
Therefore, we aim to develop a top-down visual explanation method for deep RL models.
Conventional top-down visual explanation methods are applied for supervised learning problems such as image classification.
The reason is that these methods require correct labels and the network is trained with these labels to output interpretable attention maps.
Meanwhile, deep RL networks are trained by using a reward to obtain a higher reward.
It is difficult to simply apply conventional top-down methods.

\begin{figure}[t]
\centering
\includegraphics[width=\linewidth]{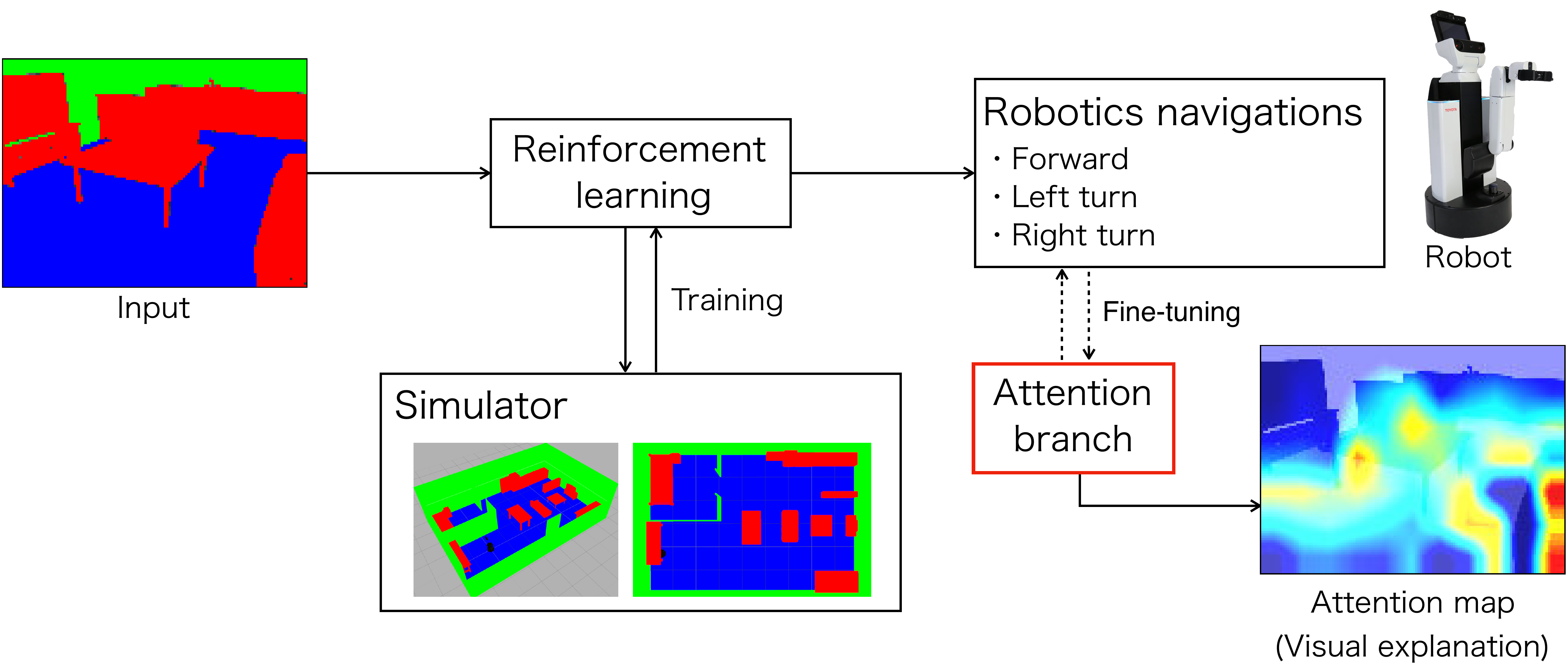}
\caption{The overview of the proposed approach. We first train a deep RL model to acquire optimal behavior for navigation task. Then, we add attention branch to the trained deep RL model and update the network parameters of attention branch. Attention map can be obtained from the trained attention branch.}
\label{fig:overview}
\end{figure}

In this paper, we propose a top-down visual explanation method for deep RL without correct labels.
The key idea of the proposed method is an attention branch network (ABN) \cite{Fukui2018} which is one of the top-down visual explanation methods.
ABN consists of the following three modules: feature extractor, attention branch, and perception branch.
The feature extractor extracts feature maps from an input image.
The extracted feature map is fed into the attention branch to obtain the attention map.
Then, the perception branch outputs the final classification score by using the attention map and the feature map.
Assuming that a deep RL model consists of a feature extractor and a perception branch, the proposed method add attention branch to the deep RL model.
This additional attention branch is trained to output attention maps corresponding to the deep RL output.
The proposed method can output the attention map with actions at the same time.
We evaluate the proposed method by using robot navigation task and demonstrate that the decision-making of deep RL model can be analyzed from attention maps.

Our contributions are as follows:
\begin{itemize}
\item We propose a top-down visual explanation method for deep RL. The proposed method does not require correct labels and we can train the network to output attention map by using actions selected by the pre-trained deep RL model.
\item The attention maps obtained by our top-down visual explanation method enables us to interpret the decision-making of the deep RL model during inference.
\end{itemize}

%===========================================================
\section{RELATED WORK}
\label{sec:related}

\subsection{Robot navigations with deep reinforcement learning}

Deep RL has been widely used for robot navigations \cite{Fangyi2015,Smolyanskiy2017,Zhu2017}.
These robot navigation methods have performed well by defining appropriate rewards that measure the distance between the start and the goal and/or time to reach the goal.
To train deep RL agents, these approaches use the real space or a simulation environment.
Training of a deep RL model with simulation environments is particularly cost-effective and is generally used.

RL models are categorized into the following three models: value-based, policy-based, and actor-critic models.
Value-based models (e.g., Q-network \cite{Tesauro1995}, deep Q-network~(DQN) \cite{Mnih2015}, and double DQN \cite{Hasselt2016}) output a Q-value $Q (\bm{a}_t | \bm{s}_t; \theta)$ that indicates the value of the action at state $s_t$.
Policy-based models (e.g., Trust Region Policy Optimization (TRPO) \cite{Schulman2015} and Proximal Policy Optimization (PPO) \cite{Schulman2017}) update an agent policy $\pi (\bm{a}_{t}|\bm{s}_{t}; \theta)$ at state $s_t$.
Actor-critic models (e.g., A3C \cite{Volodymyr2018} and Unsupervised Reinforcement and Auxiliary Learning (UNREAL) \cite{Jaderberg2016}) output both state value $V(\bm{s}_{t}; \theta)$ and policy $\pi (\bm{a}_{t}|\bm{s}_{t}; \theta)$ at state $s_t$.
Among them, we focus on the value-based models and construct a top-down visual explanation method.
The reason is that value-based RL models define discrete actions as output.
This discrete action can be considered a similar output as top-down visual explanation methods used in image classification problems including ABN.

\subsection{Visual explanations on deep neural networks}

Visual explanation methods, which analyze the reason of decision-making via attention maps, have been widely investigated in the computer vision field \cite{Zeiler2014,Zhou2016,Smilkov2017,Selvaraju2017,Fukui2018}.
The visual explanation can be categorized into two approaches; bottom-up and top-down approaches.
Bottom-up approaches generate attention maps by using response values obtained from each network layer \cite{Smilkov2017,Jost2015,Bojarski2017}.
A major bottom-up approach is SmoothGrad \cite{Smilkov2017}.
SmoothGrad generates a sensitivity map by adding noise to an input image iteratively and takes the average of these sensitivity maps.
These methods visualize attention maps by using positive gradient information for the specific class at backpropagation.
VisualBackProp \cite{Bojarski2017} has also been proposed to visualize attention maps by applying deconvolution and taking the average for feature maps obtained from each convolutional layer.
These bottom-up methods have been widely used for many applications because we can interpret various pre-trained models' decision-making without any modification and additional training of a network.

Top-down approaches generate attention maps based on the network output, especially, object classes of a classification problem \cite{Zhou2016,Selvaraju2017,Aditya2017,Fukui2018}.
In contrast to bottom-up approaches, top-down approaches require specific modification and training of a network with correct labels to generate attention maps with respect to object categories.
However, top-down approaches can output attention maps during forward propagation.
Class activation mapping (CAM) \cite{Zhou2016} is a typical top-down visual explanation method.
CAM generates attention maps of each object category by using the response of the convolutional layer and weights of a fully connected layer.
The major drawback of CAM is a decrease of classification performance because CAM needs to replace the fully connected layers near the output layer with convolutional layers.
To overcome the decrease of performance, attention branch network (ABN) \cite{Fukui2018} have been proposed.
ABN consists of three modules; feature extractor, attention branch, and perception branch.
ABN exploits the attention map obtained from attention branch for attention mechanism, which improves the classification performance.
To generate interpretable attention maps, top-down methods need to train a network by using correct labels.
Therefore, it is difficult to simply apply these methods for deep RL models.
To overcome this problem, we propose a top-down visual explanation method that can be trained without correct labels.

Visual explanations for deep RL models have also been investigated \cite{Sorokin2015,Pardo2018,Samuel2018}.
The most of visual explanations for deep RL is based on bottom-up approaches.
Samuel \textit{et al.} \cite{Samuel2018} generate a saliency map of an actor-critic model by using the network response and gradient information for perturbed images.
However, bottom-up visual explanation methods require backpropagation process as described above.
Meanwhile, we propose a top-down approach for deep RL models and the proposed method can generate attention maps during forward propagation.

%===========================================================
\section{Proposed method}
\label{sec:method}

\begin{figure*}[t]
\centering
\includegraphics[width=0.8\linewidth]{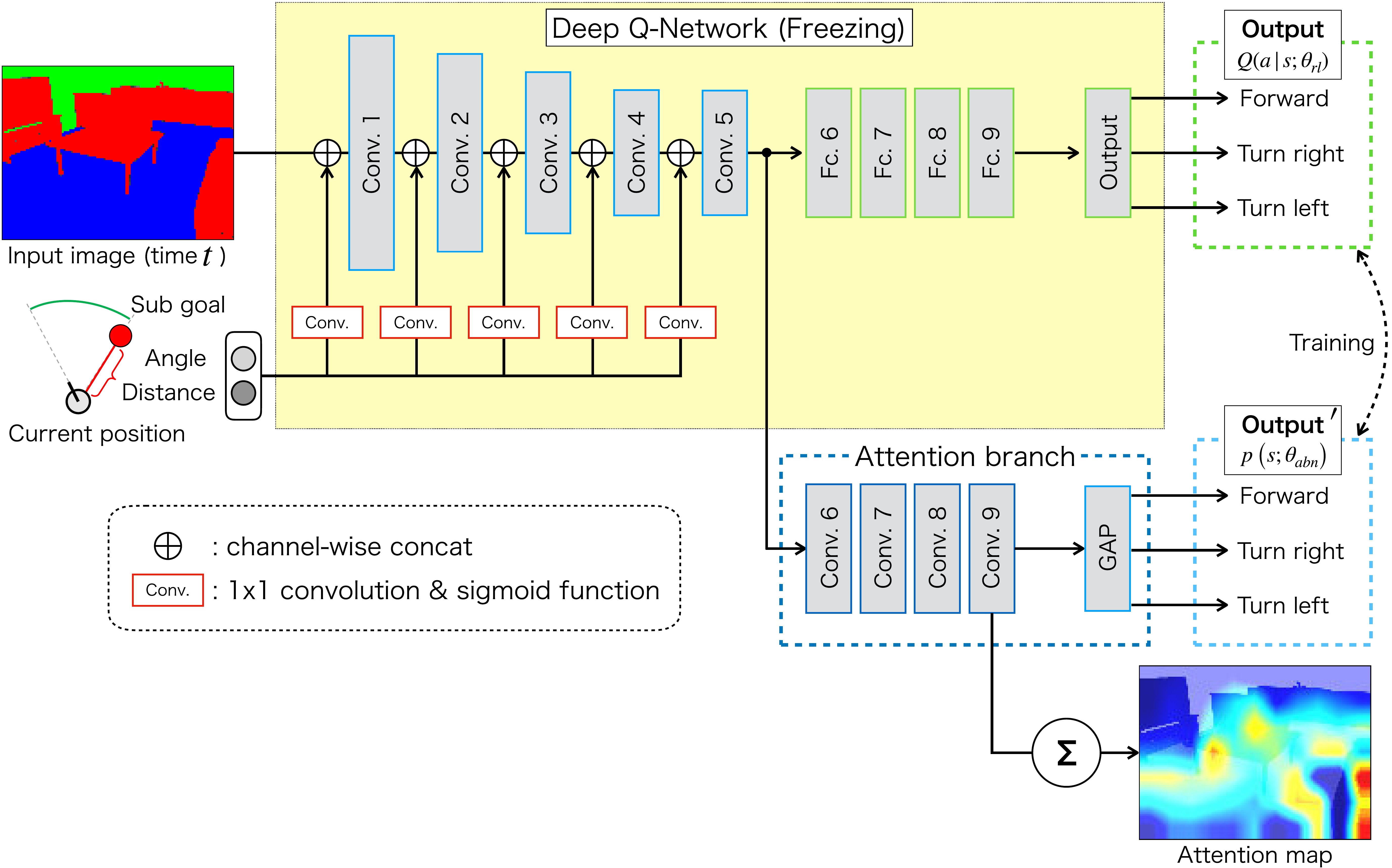}
\caption{Overview of the proposed top-down visual explanation method of deep RL model. As a deep RL model for robot navigation task, we use deep Q-network (DQN). We first train the deep Q-network to obtain higher reward and to acquire optimal behavior. We then freeze the network parameter of DQN and add an attention branch. The attention branch is trained by using output action of the trained DQN as a correct label.
For robot navigation task, the sub-goal information is embedded into network at convolutional layers so that attention map is changing respect with locations of the sub-goal.}
\label{fig:prop_abn_dqn}
\end{figure*}

In this section, we introduce the details of the proposed top-down visual explanation method for deep RL models.
However, as mentioned above, the problem to introduce top-down visual explanation method to deep RL models is the lack of correct labels because top-down method needs to be trained by using correct labels as supervised learning manner.
To overcome this problem, we utilize the structure of ABN.
ABN has two branches; attention branch that generates attention maps and perception branch that output classification result.
Focusing on this branch structure, we propose to add attention branch to a deep RL model, especially a value-based deep RL model.
We train the additional attention branch to output the same action as deep RL models.
This enables to generate attention maps that corresponding to deep RL model actions.

Figure \ref{fig:overview} shows an overview of the proposed method.
In this work, we use robot navigation with a deep RL model.
First, we train a deep RL model for robot navigation.
Then, we add an attention branch to the trained deep RL model to obtain attention maps for a visual explanation.
Hereafter, we introduce the details of i) the proposed top-down visual explanation method on deep RL with attention branch and ii) robotics navigation with deep Q-network.

\subsection{Top-down visual explanations of DQN by fine-tuning attention branch}

Figure \ref{fig:prop_abn_dqn} shows our top-down visual explanation method for deep RL models.
In this method, we use value-based deep RL models such as DQN, which consists of convolutional layers to extract feature map from the input image and fully connected layers to output Q-values for each action (the detailed structure is described in Sec. \ref{sec:dqn}).
Furthermore, we connect an attention branch behind the convolutional layers, which consists of convolutional layers and global average pooling (GAP) layer.
Attention maps are generated by taking an average of the feature maps extracted from convolutional layers in the attention branch.

To train the proposed network, it is difficult to train the entire network simultaneously in an end-to-end manner because deep RL model is trained by using rewards without correct labels.
Therefore, we train the proposed network by the following two training steps.
First, we train the DQN without the attention branch in a reinforcement learning manner to obtain higher reward and to acquire optimal action selection.
In the next step, we freeze the network parameters of DQN and update only parameters of the attention branch with the actions obtained from the trained DQN.
By using the actions output from DQN as correct labels, the attention branch is trained to output the same action selection as DQN.
Since the network parameters of DQN is frozen, the training of the attention branch does not affect to DQN performance.
Therefore, we can obtain attention maps without degrading the performance during forward propagation and can interpret the decision-making of deep RL model from the obtained attention maps.

Here, we denote the network parameter of DQN and additional attention branch as $\theta_{rl}$ and $\theta_{ab}$, respectively.
Let $s_t$ and $a_t$ be a state and action at time $t$, we define the loss function to train the attention branch $L_{ab}$ based on a cross entropy loss function as follows:
\begin{equation}
L_{ab} = B \left( Q \left(a_t | s_t ; \theta_{rl} \right) \right) \log p \left( s ; \theta_{ab} \right),
\end{equation}
where $p(s;\theta_{ab})$ is output probability from attention branch, $Q(a|s;\theta_{rl})$ is Q-value of the trained DQN, and $B(\cdot)$ is a function to make a one-hot vector from the Q-value of the DQN.
By using this loss function, attention branch is trained to select the same action as the trained DQN.

In the proposed method, we generate attention map by taking the average of feature map obtained from a convolutional layer in the attention branch.
Let $k$ be the number of channels of the feature map of the last convolutional layer in attention branch, which is corresponding to the number of actions in DQN and attention branch output.
We generate an attention map from the $k$-channels feature map by taking average over each feature map.
%
%In conventional ABN framework, attention map is generated by applying $1 \times 1 \times 1$ convolution to the $k$-channels feature map.
%However, we cannot use the $1 \times 1 \times 1$ convolution to obtain an attention map because this convolutional layer is optimized by the gradient in the perception branch.
%Therefore, in our method, we take the average over each feature maps.

\subsection{Robot navigation with deep Q-network}
\label{sec:dqn}

In this paper, we apply the proposed visual explanation method for robot navigation task for validating the effectiveness on robotic applications.
Herein, we describe the details of our deep RL model for a robot navigation task.

Firstly, as a global motion planning, we used an optimal rapidly-exploring random tree (RRT*) algorithm \cite{Karaman2011} to find a feasible global path to a given goal.
And, we set the intermediate nodes in the resultant path by RRT* as sub-goals (i.e., way points).

Then, a robot moves toward the goal by deep RL method.
We set the RL agent to reach each sub-goals by avoiding obstacles.
As we mentioned above, we use DQN as deep RL model.
Figure \ref{fig:prop_abn_dqn} shows the detailed network structure.
We input an image taking by equipped camera on a navigation robot and information of the nearest sub-goal into the network.
The sub-goal information consists of two scaler values: angle and distance.
To input such scaler values into a feature map, we make a map representing the scaler value with the same size as the feature map obtained from a convolutional layer by applying $1 \times 1$ convolution and sigmoid function.
This map is concatenated with the feature map obtained from a convolutional layer.
These additional inputs give the direction where the robot should move and facilitate network training.
The robot outputs the three actions as follows: forward, left turn, and right turn.
During the training of DQN, we update parameters by using the following reward:
\begin{equation}
\label{eq:return}
R_t =
\begin{cases}
r_{goal}  & (\textrm{reach a sub-goal}) \\
r_{dist}  & (\textrm{select forward}) \\
r_{crash} & (\textrm{crash}).
\end{cases}
\end{equation}
The reward $r_{goal}$ is given to the agent when the agent successfully reaches a sub-goal.
$r_{dist}$ is given in case that the agent selects the forward action.
The value of $r_{dist}$ is calculated by a Euclidean distance between current position and sub-goal $s$ generated by RRT* algorithm.
$r_{crash}$ is given when the agent crashes with any obstacles.

\begin{figure}[t]
\centering
\subfigure[Real: RGB image]{\includegraphics[width=0.32\linewidth]{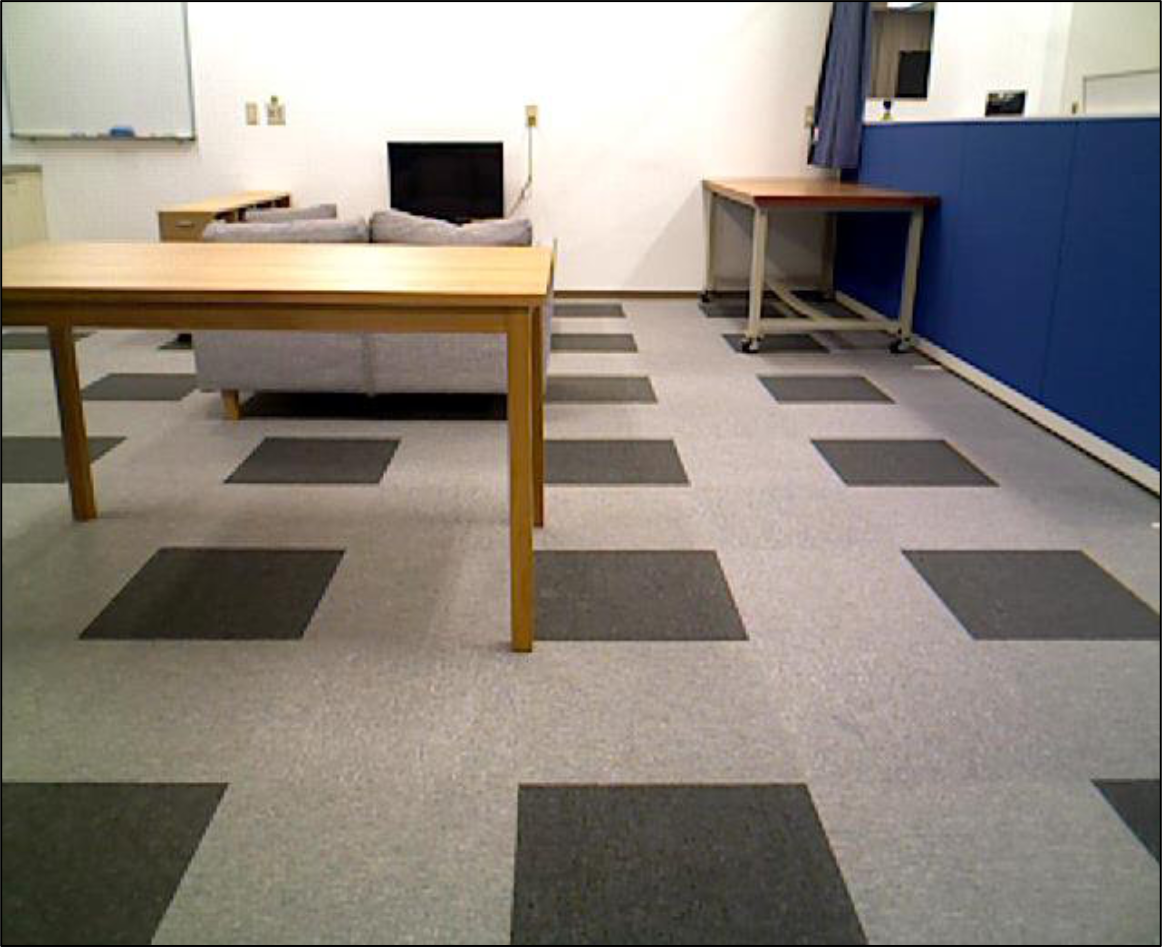}}
\subfigure[Real: depth image]{\includegraphics[width=0.32\linewidth]{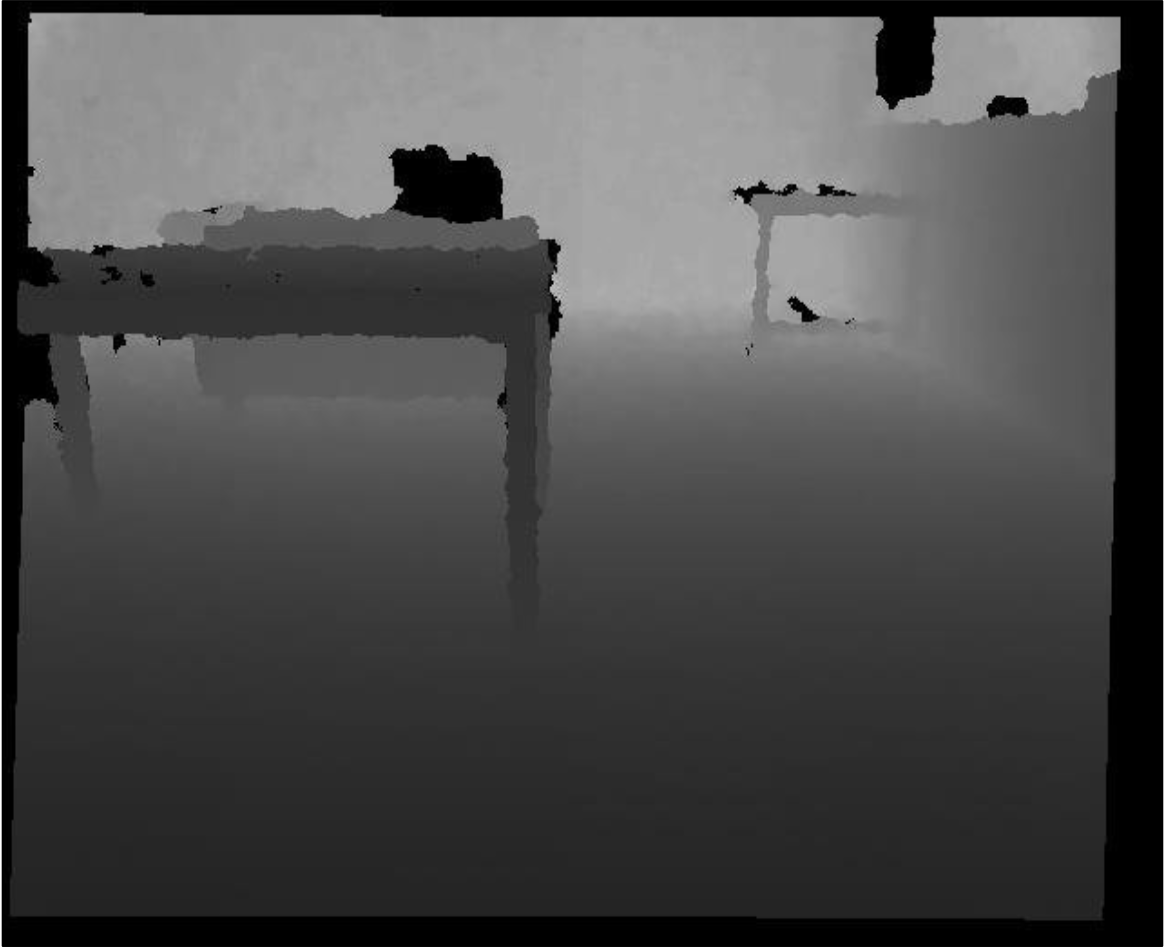}}
\subfigure[Real: semantic segmentation image]{\includegraphics[width=0.32\linewidth]{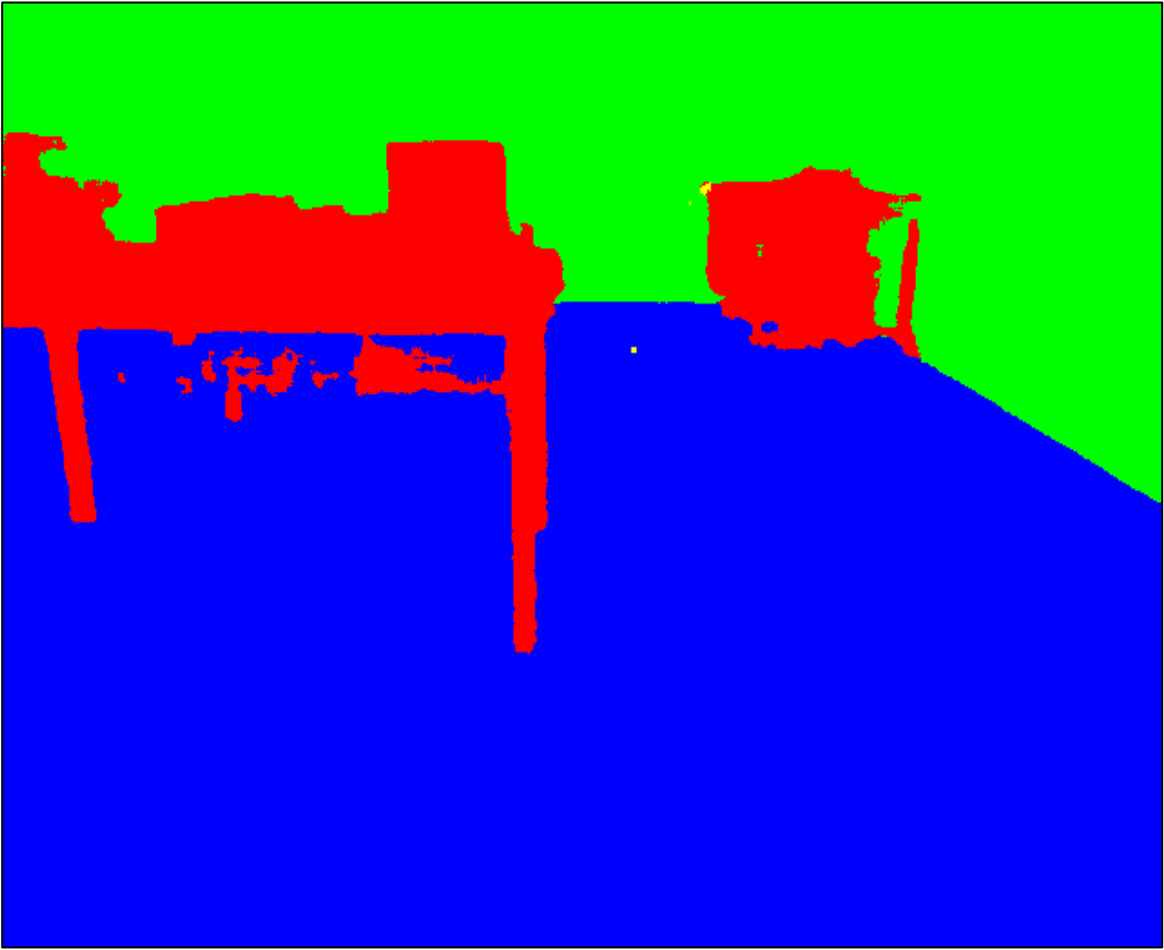}}\\
\subfigure[Simulator: RGB image]{\includegraphics[width=0.32\linewidth]{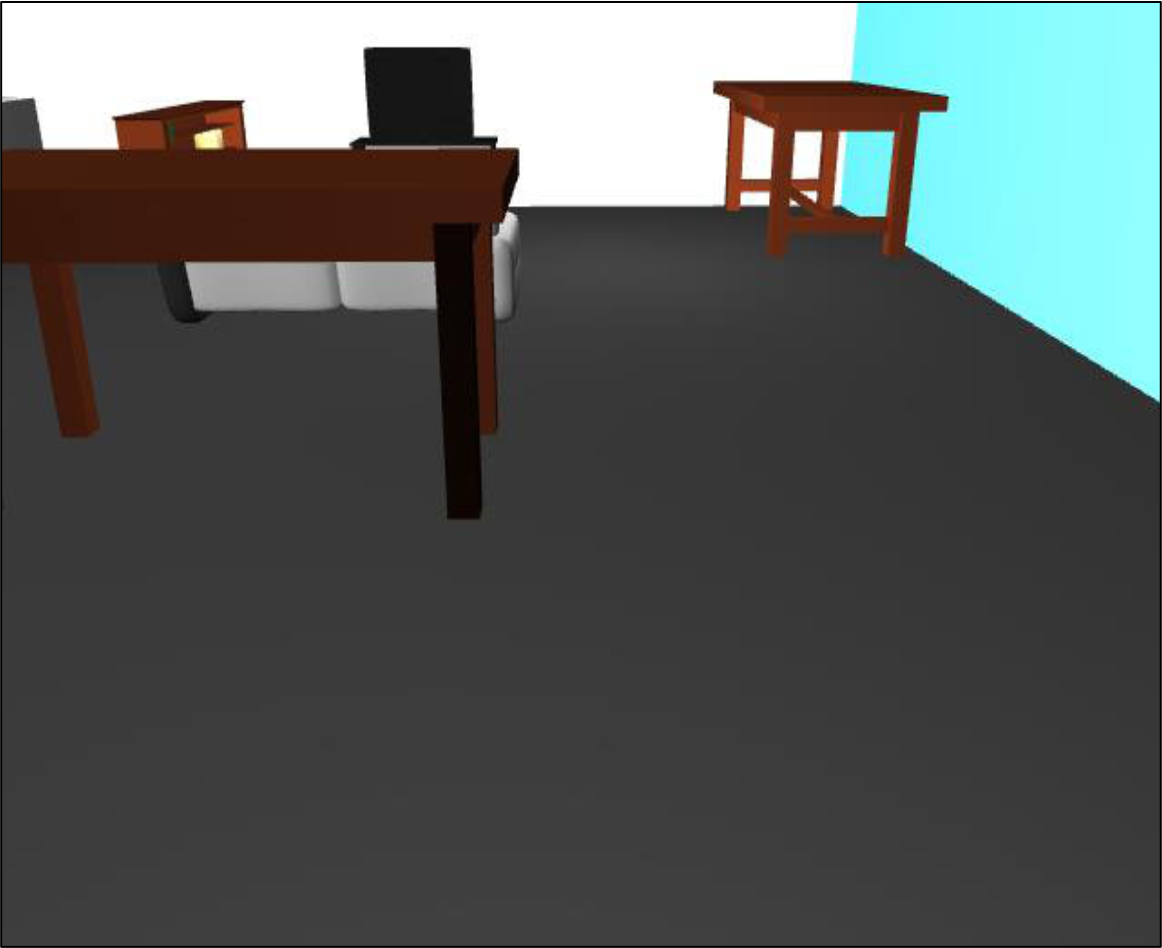}}
\subfigure[Simulator: depth image]{\includegraphics[width=0.32\linewidth]{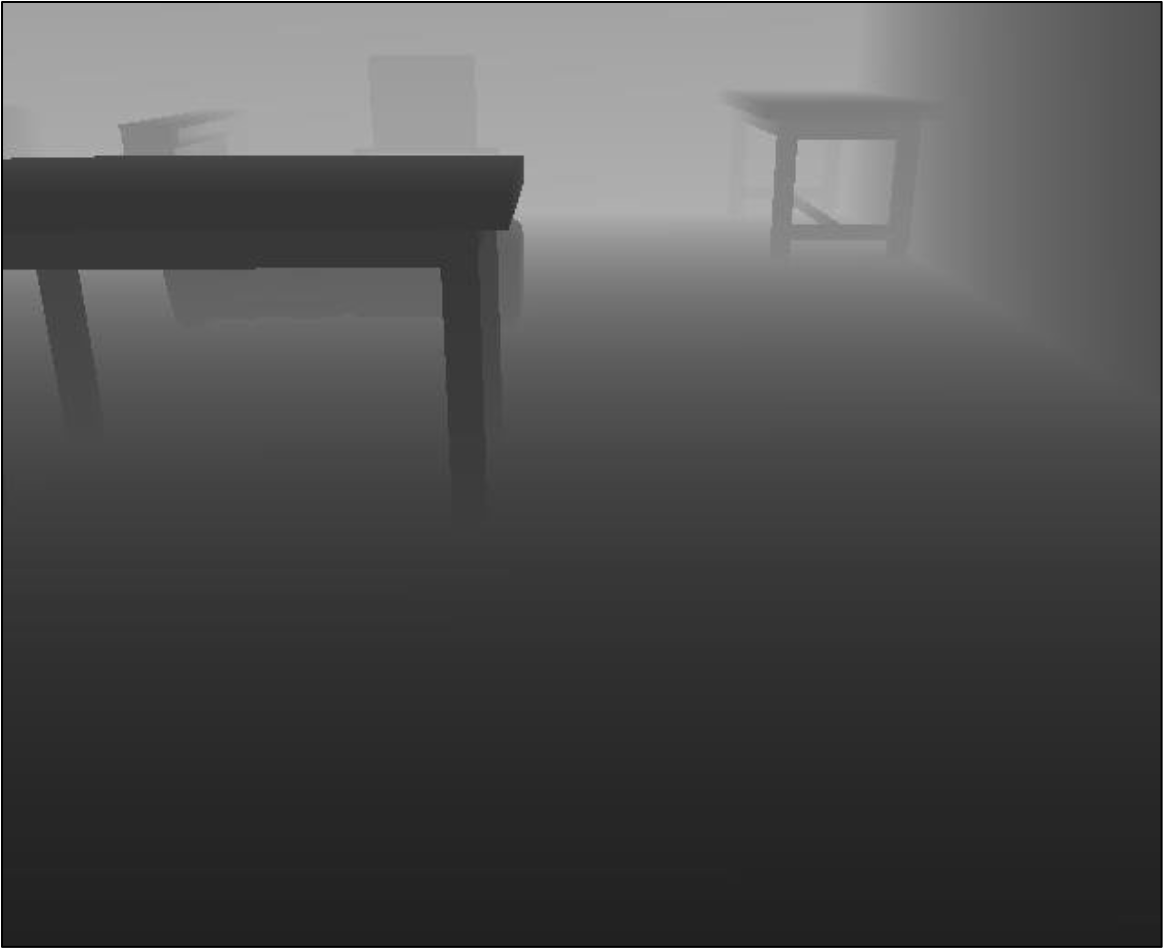}}
\subfigure[Simulator: semantic segmentation image]{\includegraphics[width=0.32\linewidth]{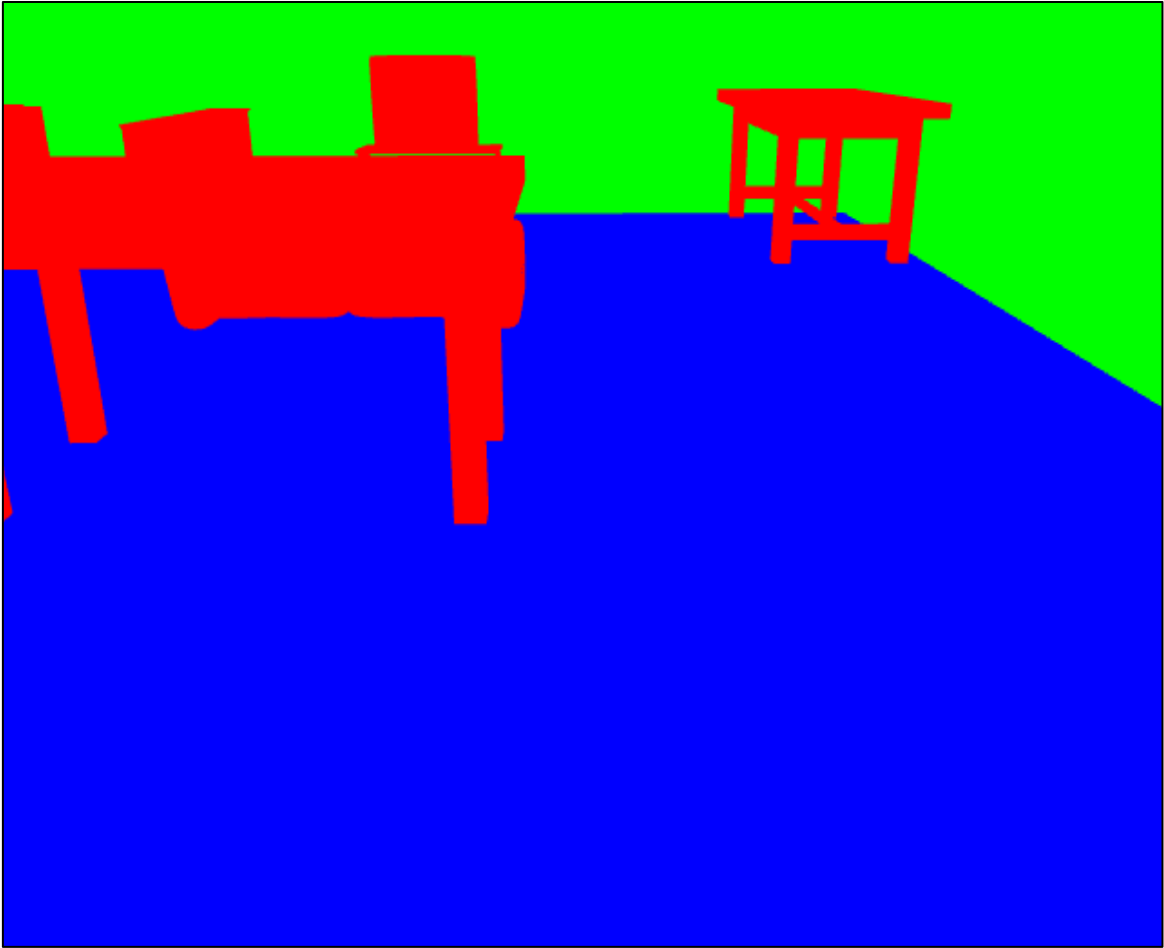}}
\caption{Examples of input images. The top row (a-c) shows images in real space and the bottom row (d-f) shows images in a simulator.}
\label{fig:example_sample}
\end{figure}

\subsection{Input image for deep RL model}
\label{sec:input}
% simulatorで評価を行う．さらに実環境でも評価を行う．そのためにSemSegを使用する．

In this paper, we use robot navigation task as an application of deep RL method and the proposed visual explanation method.
For evaluation, we use both simulator and real environments (the details are described in the experimental section).
Specifically, we train a deep RL model in a simulator environment.
Then, we evaluate the trained model in both simulator and real environments.
The reason is that the training under real environment causes the time-consuming agent training the physical damages of a robot.

As an input image into DQN, we use three kinds of image: RGB, depth, and semantic segmentation images.
Typically, robot navigation with deep RL uses raw sensor information such as RGB or depth images which are taken by front-mounted RGB-D camera on a robot as a network input and deep RL agent is trained in a simulator environment.
However, it is difficult for the deep RL agent to successfully behave in real space because of the difference of input data between a simulator and real space.
To fill the gap between a simulator and real space easily, we adopt to use semantic segmentation result as input images.
Moreover, because the use of segmentation images can give semantic information of the scene explicitly, deep RL agent can easily recognize the existence of any objects such as furniture and wall.

As a semantic segmentation method for real space, we used SegNet \cite{Vijay2015}.
Although various semantic segmentation methods have been proposed \cite{Vijay2015,Long2015,Shelhamer2017,Ronneberger2015,Hengshuang2016,Chen2018}, SegNet is developed for efficient memory and less computational time.
SegNet output segmentation result with three classes: furniture, wall, and floor.
For the network training with semantic segmentation images in the simulator, we generate segmentation images by replacing the texture of each object region with a single color.
Figure \ref{fig:example_sample} shows examples of input images in the simulator and real space.

%===========================================================
\section{Experiments}
\label{sec:experiment}

We evaluate the performance and interpretability of the proposed method with a robot navigation task.

\subsection{Experimental settings}

% table #################
%\input{table_navigation}
\begin{table*}[t]
\vspace{2truemm}
\caption{Performance of robot navigation in simulator and real environments}
\label{tab:nav_res_sim}
\centering
\tabcolsep=1.4mm
\begin{tabular}{l|cc|cc|cc}
\hline
 & \multicolumn{2}{c|}{\# of success trials} & \multicolumn{2}{c|}{Avg. dist. to goal [m]} & \multicolumn{2}{c}{\# of collisions} \\ \cline{2-7}
Input                  & Sim. & Real & Sim. & Real & Sim. & Real \\ \hline
RGB                    & 46/50 & 1/10 & 0.94 & 2.88 & 4 & 8  \\
Depth                  & 47/50 & 5/10 & 0.82 & 1.41 & 1 & 5  \\
Semantic Segmentation  & 45/50 & 8/10 & 1.12 & 0.67 & 2 & 2  \\ \hline
\end{tabular}
\end{table*}
% #######################

\begin{figure}[t]
\centering
\subfigure[]{\includegraphics[width=0.32\linewidth]{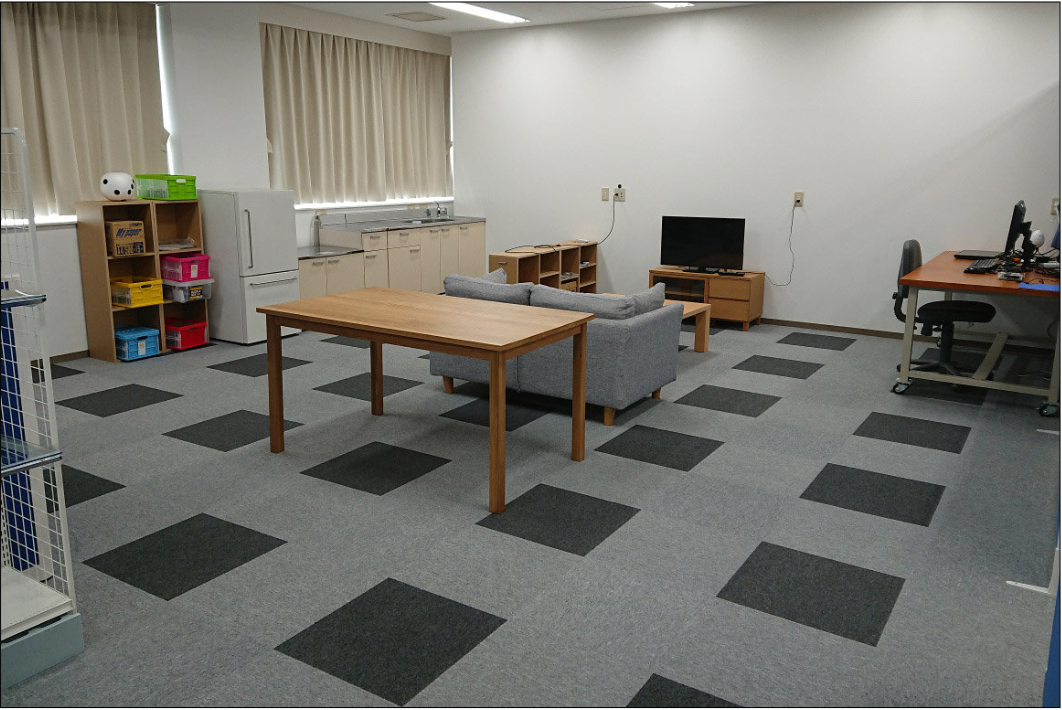}}
\subfigure[]{\includegraphics[width=0.32\linewidth]{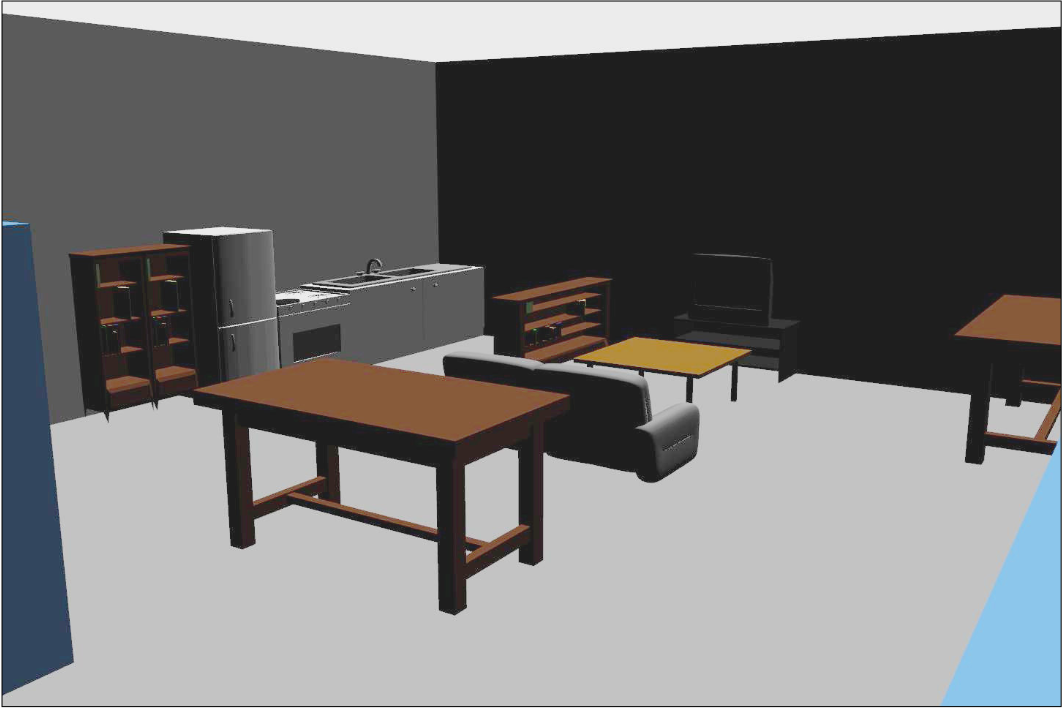}}
\subfigure[]{\includegraphics[width=0.32\linewidth]{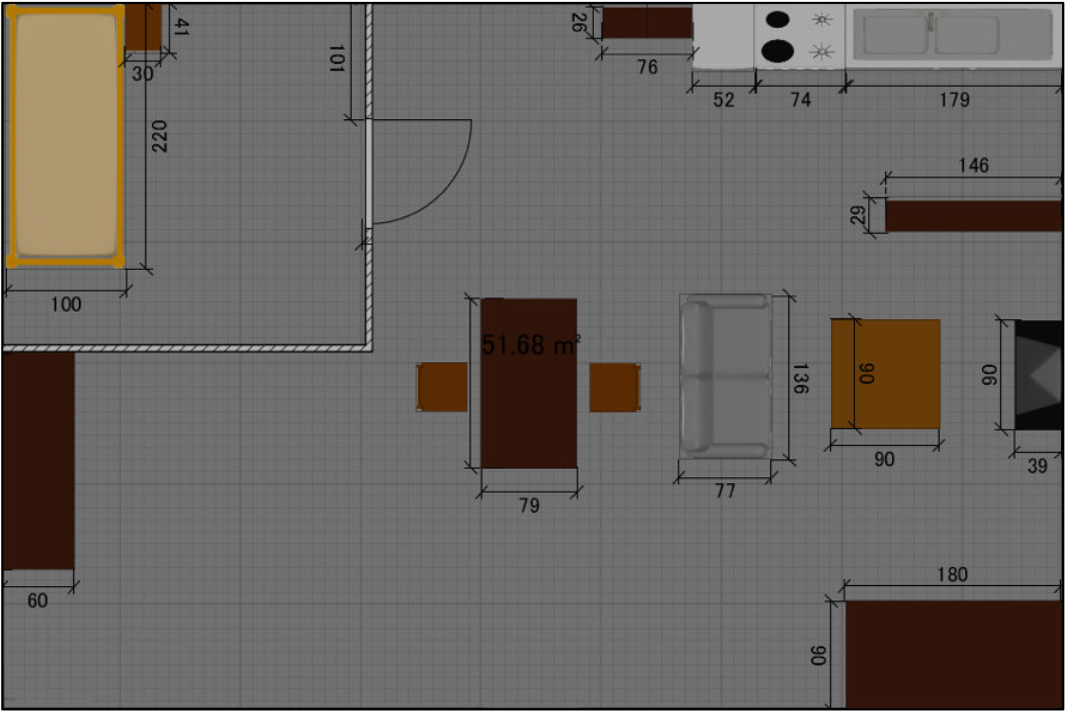}}
\caption{Our experimental environment for robot navigation. From left to right, (a) snapshot of the real space environment, (b) developed simulator corresponding to the real space, and (c) a map of the environment.}
\label{fig:env_map}
\end{figure}

\subsubsection{Experimental environment}
For robot navigation task, we used human support robot (HSR).
Figure \ref{fig:env_map} shows our experimental environment.
We prepared a room that imitates a human's living space and developed a simulator environment that is the same as the real space.

\subsubsection{DQN training}
During the training of the DQN and attention branch, we train the DQN agent by using the developed simulator.

For the training of DQN, we define the rewards $r_{goal}$ and $r_{crash}$ in Eq. (\ref{eq:return}) as $30$ and $-5$, respectively.
Also, we set the discount factor $\gamma$ as 0.99 and the size of replay buffer for experience replay as $1.0 \times 10^4$.
And, we used linear decay $\epsilon$-greedy for exploration.
The $\epsilon$ is linearly decreased from 0.1 to 0.9 until $8.0\times10^{4}$ episodes.

For the training of attention branch, we update the parameters 100 epochs by using stochastic gradient descent (SGD) with momentum.
The learning rate is initially set as 0.1 and is divided by 10 at 50 epochs and 75 epochs.

\subsubsection{Semantic segmentation}
For the semantic segmentation of real space images, we collected 652 images and we manually annotated labels.
We annotated by using the following three classes: furniture, wall, and floor.
During the training, to increase the number of training samples, we randomly applied data augmentations such as horizontal flip, crop, contrast conversion, or add noise to training samples.

\subsection{Performance of robot navigation}
\label{sub:attention_analysis}

Before demonstrating and evaluating the proposed visual explanations method, we first evaluate the performance of conducted robot navigation task.

As we mentioned before, we conduct the navigation task in both simulator and real environments.
For the evaluation in the simulator, we randomly initialize the starting point of a robot and give the goal coordinate.
Then, the deep RL agent moves to reach goals.
For the evaluation in the real environment, we manually place a robot and set a goal coordinate.
Then, a robot moves toward the goal based on the trained deep RL model decisions.
We repeat this trial with changing start and goal, and we evaluate how many the agent successfully reaches goals.
We tried this trial 50 times for simulator environment and 10 times for real environment.

We used the following three evaluation metrics: the number of trials to reach the goal, the average distance to goal coordinate at the end of trials, and the number of collisions.
In order to decide if the robot reach the goal, we measure the distance between the current position and the goal.
In case the distance is lower than 0.5 meters, we assume that the robot reaches the goal.

Table \ref{tab:nav_res_sim} shows the evaluation results with each input image type in both simulator and real environments.
The results of simulator environment show that each input image type successes to reach goals.
Among them, the result with semantic segmentation image provides rather poorer results than the other image types.
Meanwhile, with respect to the real environment, the results show that RGB and depth image types failed to reach the goals.
The reason for the failure is the difference of domains between a simulator and real space.
Especially, as shown in Fig. \ref{fig:example_sample}, RGB images are greatly different.
In fact, the number of success trials also decreases.
In contrast, the deep RL agent with semantic segmentation stably reaches the goal even if the agent is trained in a simulator.

In the following evaluation of the proposed visual explanation method and demo results in a real environment, we use semantic segmentation image as an input for deep RL model.

\begin{figure}[t]
\centering
\includegraphics[width=\linewidth]{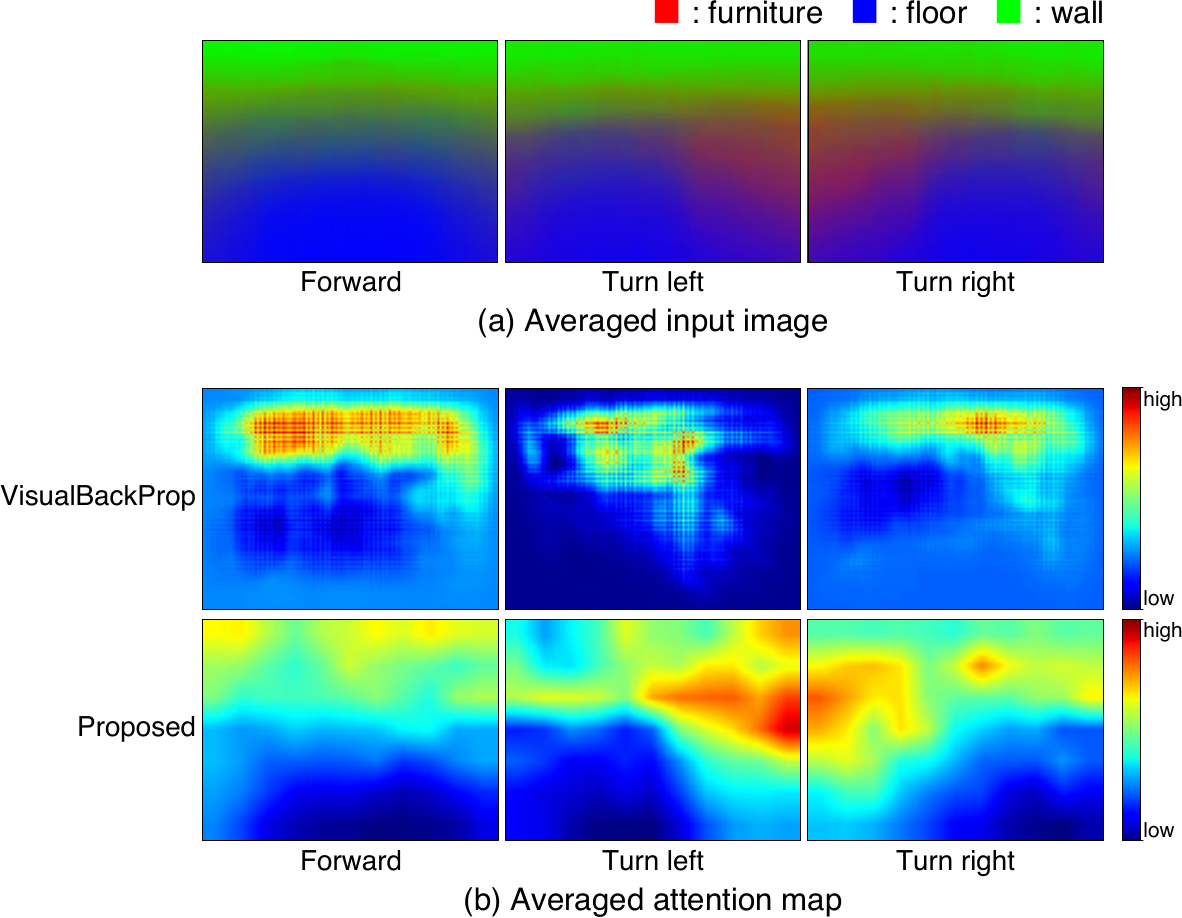}
\caption{(a) Averaged input images and (b) attention maps for each action.}
\label{fig:ave_attention}
\end{figure}

\subsection{Analysis of attention maps}

To analyze the relationship between the selected actions and the obtained attention maps, we calculate the averaged attention maps for each action.
Figure \ref{fig:ave_attention} shows the averaged images and attention maps with respect to selected actions.
The left of Fig. \ref{fig:ave_attention} shows the result when an agent selects forward action and the attention map highlights the front of an agent uniformly.
In contrast, attention maps for turn left or right actions highlight the region that corresponds to the selected actions, respectively.
Moreover, the averaged input images of left and right turn contain red regions at the same regions where the attention maps are highly responded.
These results imply that the agent focuses on the direction of the next movement and check the existence of obstacles.
On the other hand, the attention maps of VisualBackProp for left and right turns highlight rather the center of images also that of forward action as well.
The attention map of VisualBackProp has no significant difference between each action and VisualBackProp has lower interpretability.

\begin{figure}[t]
\centering
\includegraphics[width=\linewidth]{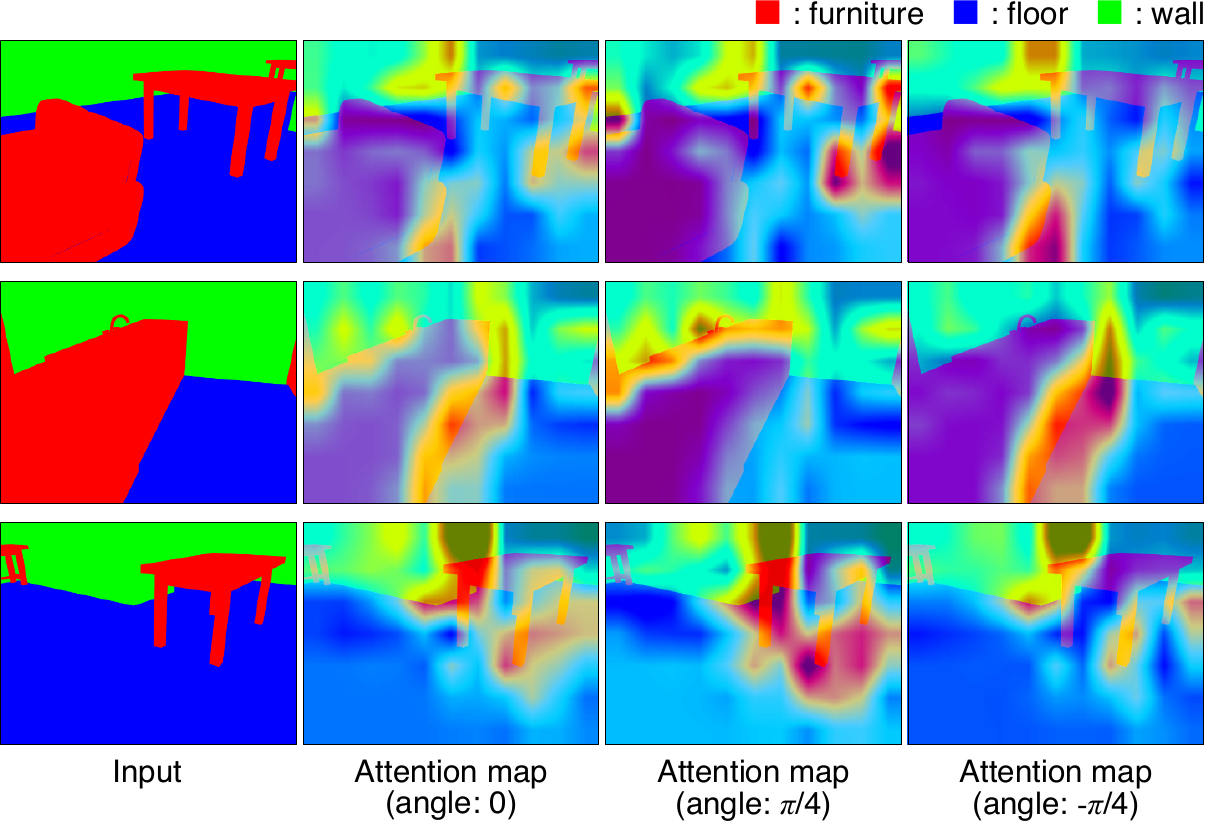}
\caption{Attention maps over different input angles. From left to right, the input semantic segmentation image, attention maps with sub-goal information of front facing direction (angle is zero), right direction (angle is $\pi / 4$), and left direction (angle is $- \pi / 4$).}
\label{fig:comp_angle}
\end{figure}

Next, we analyze the relationship between the sub-goal input and an attention map.
To compare the difference with respect to the sub-goal input, we artificially input different angles for the same image.
The used angle is $0$ (front), $\pi/4$ (right front), and $-\pi/4$ (left front) [rad].
The distance to the sub-goal is set as 1.0 [m].
Figure \ref{fig:comp_angle} shows the obtained attention maps with different input angles.
These results show that the focused regions are changed depending on the input angles.
For example, in the case of the top row of Fig. \ref{fig:comp_angle}, attention map of $\pi/4$ highly focuses on furniture on the right side of the image while that of $-\pi/4$ highlights furniture on the left side of the image.
Surprisingly, the attention map of angle $0$ seems to be the average of the attention maps of $\pi/4$ and $-\pi/4$.
From these result, the network considers the existence of obstacles and the direction of movement.

\begin{figure}[t]
\centering
\subfigure[Deletion]{
\includegraphics[width=0.7\linewidth]{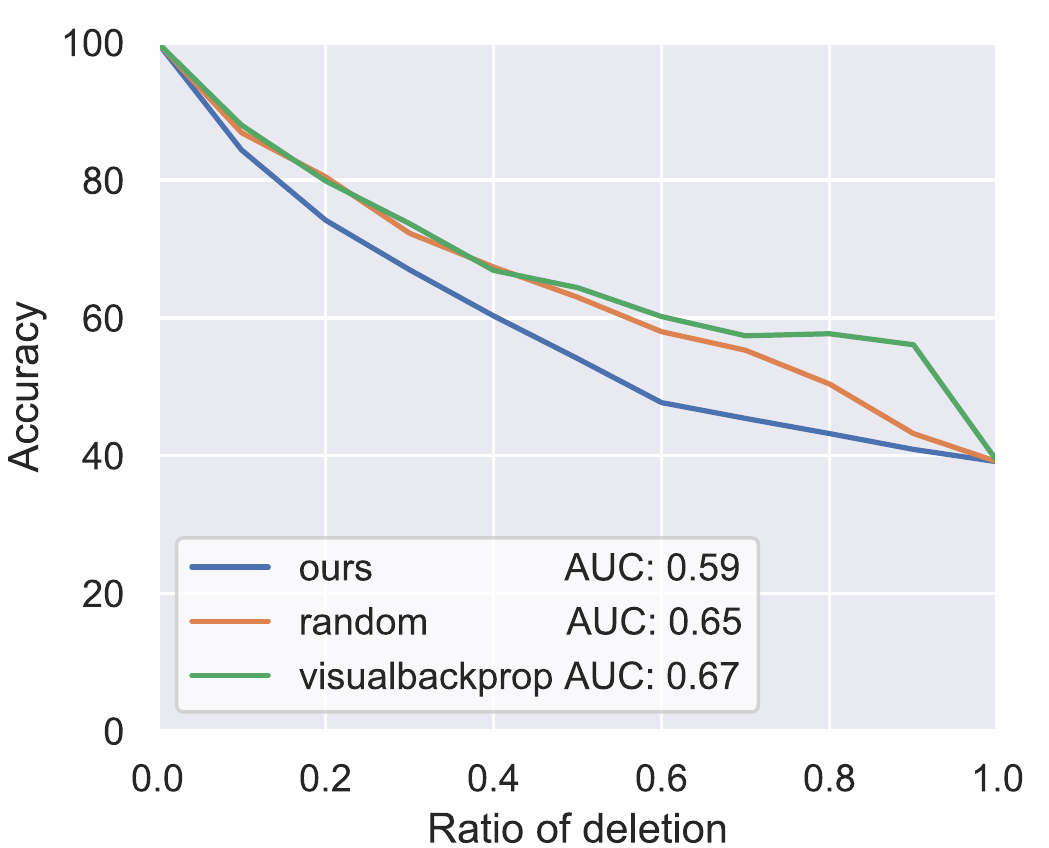}}
\subfigure[Insertion]{
\includegraphics[width=0.7\linewidth]{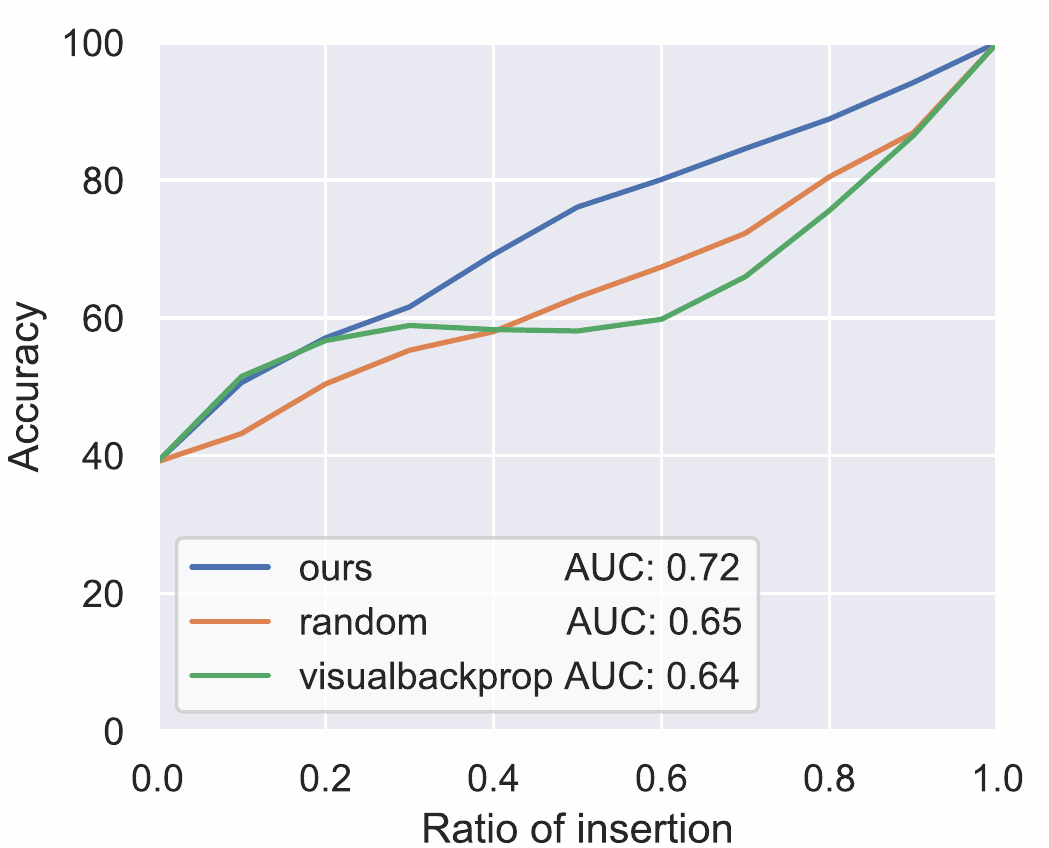}}
\caption{Quantitative evaluation of (a) deletion and (b) insertion. In case of deletion, lower accuracy indicates the better explainability of a model. In contrast, in case of insertion, higher accuracy indicates the better explainability of the model.}
\label{fig:deletion_insetion}
\end{figure}

\subsection{Quantitative evaluation of attention map}

In order to quantitatively evaluate the explainability of the attention map, we use the deletion metric, the insertion metric \cite{Petsiuk2018rise}.
These metrics are used for evaluating explainability of attention map in image classification tasks.
The deletion metric measures the decrease of classification accuracy by gradually deleting the high attention area of an attention map from the input image.
Therefore, drastic accuracy decrease means attention area is correctly corresponding to important region for the network decision and indicates the explainability of the attention map is high.
On the other hand, the insertion metric measures the increase of accuracy by gradually inserting the high attention area of an attention map in the input image.
Therefore, drastic accuracy increase means a higher explainability.

In our case, we assume the output from DQN as the correct label and we evaluate the deletion and insertion scores of attention branch.
We compare the deletion and insertion metrics with following two baselines.
One is \textit{random}, which randomly decide the region of deletion and insertion.
The other is \textit{VisualBackProp} \cite{Bojarski2016}, which is one of bottom-up visual explanation method.
VisualBackProp computes an attention map by merging feature maps obtained from each convolutional layer in a network.

We show the deletion (lower is better) and insertion (higher is better) scores of the proposed methods in Fig. \ref{fig:deletion_insetion}.
As shown in these results, the deletion of the proposed method is lower than those of random and VisualBackProp, and the insertion of the proposed method is higher.
Consequently, the proposed method provides the better visual explanation of DQN.

\begin{figure*}[t]
\centering
\includegraphics[width=0.65\linewidth]{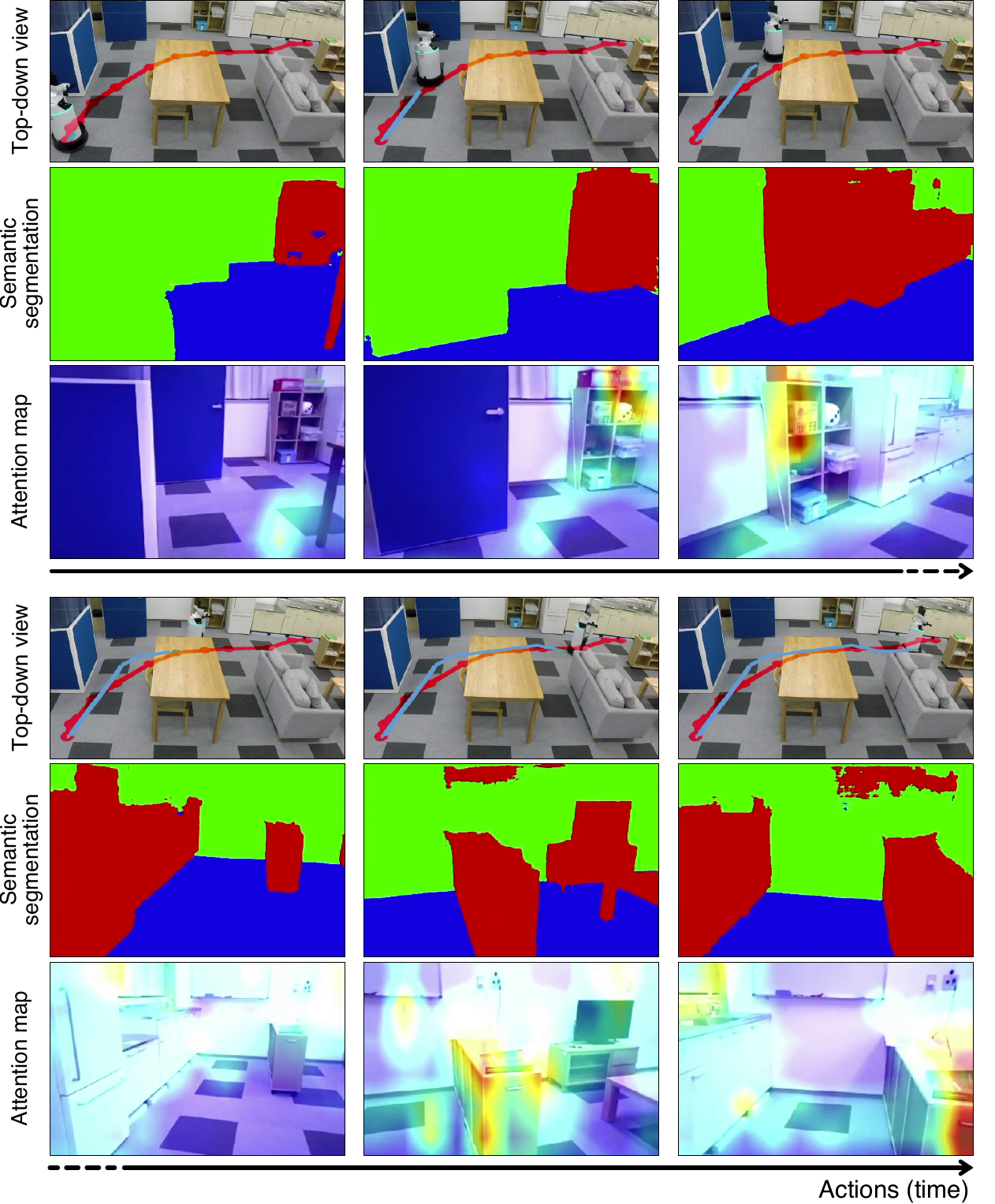}
\caption{Snapshots of our robot navigation experiment in a real environment. From top to bottom for each row, we show a top-down view of navigation experiment in a real environment, semantic segmentation results obtained by a SegNet that are input to the trained DQN, and the corresponding attention maps obtained from the proposed method. Note that the original video can be found in our supplementary material.}
\label{fig:demo}
\end{figure*}

\subsection{Demo results of robot navigation in a real environment}

At the end of our navigation experiments, we show the navigation results in the real environment.
Figure \ref{fig:demo} shows the snapshot of a demo movie of robot navigation.
The reader can watch the video in our supplementary material.
Although semantic segmentation predicted by SegNet contains wrong classification regions partially, the proposed method (i.e., DQN) successfully navigates a robot to a goal.
Note that we train the network model of the proposed method in a simulator environment.
The obtained attention map mainly highlights furniture.
However, if the robot can move forward without any collisions (e.g., the bottom-left image in Fig. \ref{fig:demo}), the attention map is not highlighted so much higher values.
From these results, we can understand the reason for the decision making of a deep RL model.

\section{Conclusion}
\label{sec:conclusion}

In this paper, we proposed a top-down visual explanation method for deep RL.
The proposed method is based on the key structure of ABN and we connect the attention branch with DQN.
The connected attention branch is trained by using the output action of DQN as a correct label in a supervised learning manner.
We evaluate the interpretability of the attention map obtained by the proposed method by using robot navigation task.
The experimental results show that the obtained attention maps highlight obstacles such as furniture and boundaries between wall and floor.
These results indicate that the used DQN model select action while paying attention to obstacles.

Although we adopted a rather simple robot navigation task for the ease of analysis of interpretability, our visual explanation method has much potential to apply various deep RL models and problems, which is one of our future work.
Moreover, our future works include developing top-down visual explanation method for another deep RL methods, i.e., policy-based and actor-critic models, and linguistic explanation.

%%%%%%%%%%%%%%%%%%%%%%%%%%%%%%%%%%%%%%%%%%%%%%%%%%%%%%%%%%%%%%%%%%%%%%%%%%%%%%%%
\section*{ACKNOWLEDGEMENT}
This paper is based on results obtained from a project, JPNP20006, commissioned by the New Energy and Industrial Technology Development Organization (NEDO).



% \section*{ACKNOWLEDGMENT}

% The preferred spelling of the word ÒacknowledgmentÓ in America is without an ÒeÓ after the ÒgÓ. Avoid the stilted expression, ÒOne of us (R. B. G.) thanks . . .Ó  Instead, try ÒR. B. G. thanksÓ. Put sponsor acknowledgments in the unnumbered footnote on the first page.

%%%%%%%%%%%%%%%%%%%%%%%%%%%%%%%%%%%%%%%%%%%%%%%%%%%%%%%%%%%%%%%%%%%%%%%%%%%%%%%%
%\bibliographystyle{IEEEtran}
%\bibliography{references}

\end{document}